\documentclass[journal,10pt]{IEEEtran}

\usepackage{epsfig}
\usepackage{graphicx}
\usepackage{caption}
\usepackage{amsmath,amsthm}
\usepackage{amssymb}
\usepackage{mathrsfs} 
\usepackage{enumerate}
\usepackage{multirow}
\usepackage{tabularx}
\usepackage{float}
\usepackage[numbers,sort]{natbib}
\usepackage{color}
\usepackage{times}

\usepackage[breaklinks=true,bookmarks=false]{hyperref}

\def \x{\mathbf{x}}

\def\h{\mathbf{h}}
\def\f{\mathbf{f}}
\def\g{\mathbf{g}}

\def\c{\mathbf{c}}
\def\Ih{\hat{I}}
\def\It{\tilde{I}}
\def\R{\mathbb{R}}
\def\C{\mathbb{C}}
\def\P{\mathbb{P}}
\def\Q{\mathbb{Q}}
\def\O{\mathcal{O}}
\def\Rad{\mathscr{R}}
\def\iRad{\mathscr{R}^{-1}}
\def\F{\mathscr{F}}
\def\iF{\mathscr{F}^{-1}}
\def\brho{\boldsymbol{\rho}}
\def\btau{\boldsymbol{\tau}}
\def\bmu{\boldsymbol{\mu}}

\usepackage{setspace}

\begin{document}

\title{The Radon cumulative distribution transform and its application to image classification}

\author{\IEEEauthorblockN{Soheil Kolouri\IEEEauthorrefmark{1}, 
Se Rim Park\IEEEauthorrefmark{2},
and Gustavo K. Rohde\IEEEauthorrefmark{1}\IEEEauthorrefmark{2}\IEEEauthorrefmark{3}}\\
\IEEEauthorblockA{\IEEEauthorrefmark{1}{
Department of Biomedical Engineering, Carnegie Mellon University, Pittsburgh, PA, 15213}\\
\IEEEauthorrefmark{2}{
Department of Electrical and Computer Engineering, Carnegie Mellon University, Pittsburgh, PA, 15213}\\
\IEEEauthorrefmark{3}{
Lane Center for Computational Biology, Carnegie Mellon University, Pittsburgh, PA, 15213}
} 
\\ {\tt\small skolouri@andrew.cmu.edu}
}
\markboth{}%
{Shell \MakeLowercase{\textit{et al.}}: Bare Demo of IEEEtran.cls for Journals}

\maketitle
\IEEEpeerreviewmaketitle
\begin{abstract}
Invertible image representation methods (transforms) are routinely employed as low-level image processing operations based on which feature extraction and recognition algorithms are developed. Most transforms in current use (e.g. Fourier, Wavelet, etc.) are linear transforms, and, by themselves, are unable to substantially simplify the representation of image classes for classification. Here we describe a nonlinear, invertible, low-level image processing transform based on combining the well known Radon transform for image data, and the 1D Cumulative Distribution Transform proposed earlier. We describe a few of the properties of this new transform, and with both theoretical and experimental results  show that it can often render certain problems linearly separable in transform space.

\end{abstract}

\section{Introduction}
Image pattern recognition is an important problem in wide variety of disciplines including  computer vision, image processing, biometrics, and remote sensing.   The primary goal of pattern recognition is supervised or unsupervised classification of data points (e.g. signals or images). Image transforms have long been used as low level representation models to facilitate pattern recognition by simplifying feature extraction from images. The Fourier transform, Walsh-Hadamard transform, wavelet transform, ridgelet and curvelet transforms, sine and cosine transforms, Radon transform, etc. are examples of such image transforms.

Some interesting applications of invertible image transforms in pattern recognition are presented in \cite{li2002palmprint,monro2007dct,do2002wavelet,mandal2007face, boulgouris2007gait}. In \cite{li2002palmprint}, discrete Fourier transform (DFT) was used for palm print identification. Monro et al. \cite{monro2007dct} used discrete cosine transform (DCT) for iris recognition. Wavelet coefficients were used as texture features in \cite{do2002wavelet} for image retrieval. Mandal et al. \cite{mandal2007face} used curvelet-based features for face recognition. The Radon transform was used for Gait recognition in \cite{boulgouris2007gait}. The list above is obviously not exhaustive. They represent just but a few examples of many applications of image transforms in pattern recognition.

\begin{figure}[t]
\centering
\includegraphics[width=\columnwidth]{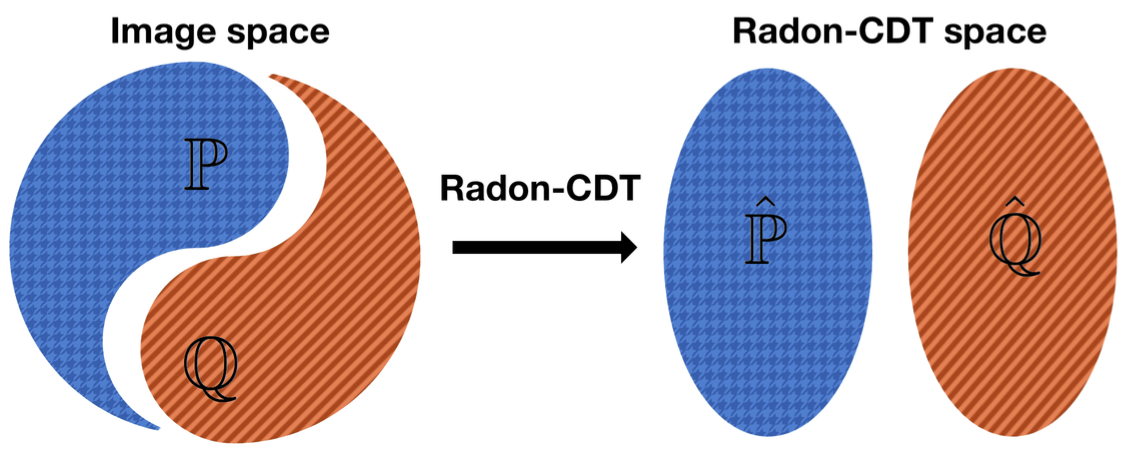}
\caption{Overview of role of Radon-CDT in enhancing linear separations of classes.}
\label{fig:classes_intro}
\end{figure}

A common property among aforementioned transforms is that they are all invertible linear transforms that seek to represent a given image as a linear combination of a set of functions (or discrete vectors for digital signals). What we mean by an invertible linear transform, $\F$, is that for images $I$ and $J$, $\F$ satisfies $\F(I)+\F(J)=\F(I+J)$, $\F(\alpha I)=\alpha \F(I)$, and $\iF$ exists. Linear transforms are unable to alter the `shape' of image classes (i.e. distribution of the point cloud data) so as to fundamentally simplify the actual classification task. For example, linear operations are unable to render classification problems that are not linearly separable into linearly separable ones (see Figure \ref{fig:classes_intro}). When considering many important image classification tasks, it is not hard to understand the problem at an intuitive level. One can often visually observe that in many image categories (e.g. human faces, cell nuclei, galaxies, etc.) a common way in which images differ from one another is not only in their intensities, but also in where the intensities are positioned. By definition, however, linear image transforms must operate at fixed pixel coordinates. As such, they are unable to move or dislocate pixel intensities in any way. Hence, for pattern recognition purposes, linear image transforms are usually followed by a nonlinear operator to demonstrate an overall nonlinear effect (e.g. thresholding in curvelet and wavelet transforms, magnitude of Fourier coefficients, blob detection/analysis in Radon transform, etc.).  

Many  feature extraction methods have been developed for images \cite{chan2014pcanet,zhang2014learning,shao2014feature,zhu2014weakly} along side with the end to end deep neural network approaches such as convolutional neural networks (ConvNets) \cite{lecun1998gradient,krizhevsky2012imagenet} and scattering networks (ScatNets) \cite{bruna2013invariant, sifre2013rotation,zhu2014weakly}. These recent methods have proven to be very successful in image classification and they have improved the state of the art classification for a wide range of image datasets. Such methods, however, are often not well suited for image modeling applications, including imaging and image reconstruction, as they provide a noninvertible nonlinear mapping from the image space to the feature space. Meaning that while the nonlinearity of the image classes are captured through the extracted features, any statistical analysis in the feature space does not have a direct interpretation in the image space as the mapping is noninvertible.

Intensity vector flows can represent an interesting alternative for encoding the pixel intensity movements which may help simplify certain pattern recognition tasks. In earlier work  \cite{wang2013linear,kolouri2015continuous} we have described a framework that makes use of the $L_2$ optimal transport metric (Earth Mover's distance) to define a new invertible image transform. The framework makes use of a reference (template) image to which a given image is morphed using the optimal transport metric. The optimal transport provides a unique vector flow that morphs the input image into the reference image. The mapping from an image to its designated vector flow can be thought of as a nonlinear image transform. Such a transform is invertible and the inverse transform is obtained by applying the inverse of the computed flow to the established template. Thus, the technique can be used to extract information regarding the pixel intensities, as well as their locations (relative to the reference image). The approach has proven useful in a variety of applications  \cite{wang2013linear,basu2014detecting,kolouri2015continuous,kolouri2015transport}. In \cite{kolouri2015transport} we have shown that it can be used to design powerful solutions to learning based inverse problems. We've also shown that encoding pixel movements as well as intensities can be highly beneficial for cancer detection from histopathology \cite{ozolek2014accurate} and cytology \cite{tosun2015detection} images. In addition, given that the transform is invertible, the approach enables visualization of any regression applied in transform space. It thus enables one to visualize variations in texture and shapes \cite{wang2013linear, basu2014detecting}, as well as to visualize discriminant information by `inverting' classifiers \cite{wang2013linear}. 

The transport-based approach outlined above, however, depends on obtaining a unique transport map that encodes an image via minimization of a transport metric. This can be done via linear programming for images that can be considered as discrete measures \cite{wang2011optimal}, or via variational minimization for images that can be considered as smooth probability densities \cite{haker2004optimal}. It is thus relatively cumbersome and slow for large images. Moreover, the mathematical analysis of any benefits regarding enhanced classification accuracy is difficult to perform given the underlying (nonlinear) minimization problem.


In this paper we describe a new 2D image transform by combining the standard 2D Radon transform of an image with the 1D Cumulative Distribution Transform (CDT) proposed earlier \cite{serim2015}. As with our earlier work \cite{wang2013linear, serim2015}, the transform utilizes a reference (or template), but in contrast to our earlier work, it can be computed with a (nonlinear) closed form formula without the need for a numerical minimization method. An added benefit of this framework is that several of its properties (including enhancements in linear separation) can now be shown mathematically. We show theoretically and experimentally,  that the newly defined Radon-CDT improves the linear separability of image classes.

 We note that, the Radon transform has been extensively used in imaging applications such as Computerized Tomography (CT) and Synthetic Aperture Radar (SAR) \cite{natterer1986mathematics, patel2010compressed}. In addition there has been a large body of work on utilizing the Radon transform to design image features that have invariant properties \cite{jafari2005radon,dahyot2013generalised,dahyot2009statistical}. What differentiates our work from the invariant feature extraction methods that also use the Radon transform is: 1) the Radon-CDT is a nonlinear and invertible image transform that enables any statistical analysis in the transform space to be directly inverted to the image space, and 2) we provide a theorem that guarantees linear separation of certain image classes in the transform space. 

 In what follows, we start by briefly reviewing the concept of the cumulative distribution transform \cite{serim2015} and the Radon transform.  In Section \ref{sec:Radon-CDT}, we introduce the Radon cumulative  transform and enumerate some of its properties. The details for the numerical implementation of our method is presented in \ref{sec:numerics}. In Section \ref{sec:results} we demonstrate the capability of the Radon-CDT to enhance linear separability of the data classes on synthetic and real-life image datasets. Finally, we conclude our work in Section \ref{sec:con}.

\section{Preliminaries}

We start by reviewing definitions and certain basic properties of the cumulative distribution \cite{serim2015} and Radon transforms.

\subsection{The Cumulative Distribution Transform}
The CDT \cite{serim2015} is a bijective nonlinear signal transform from the space of smooth probability densities to the space of differentiable functions.  In contrast to linear signal transformation frameworks (e.g. Fourier and Wavelet transforms) which only employ signal intensities at fixed coordinate points, thus adopting an `Eulerian' point of view (in PDE parlance), the idea behind the CDT is to consider the intensity variations together with the locations of the intensity variations in the signal. Therefore, the CDT adopts a `Lagrangian' point of view (in PDE parlance) for analyzing signals. 

More formally, let $\mu$ and $\sigma$ be two continuous probability measures on $\R$ with corresponding positive densities $I$ and $I_0$, such that $\int_\R d\mu(x)=\int_\R I(x)dx=1$ and $\int_\R d\sigma(x)=\int_\R I_0(x)dx=1$. The forward and inverse CDT transform (analysis and synthesis equations) of $I$ with respect to $I_0$ are defined as \cite{serim2015}, 
\begin{eqnarray}
\left\{ \begin{array}{l}
\It=(f-id)\sqrt{I_0} \\
I= (f^{-1})' (I_0\circ f^{-1})
\end{array}
\right.
\end{eqnarray}
where $(I_0 \circ f)(x) = I_0(f(x))$,  $id:\R\rightarrow\R$ is the identity function, $id(x)=x,~\forall x\in\R$, and $f:\R\rightarrow\R$ is a measurable map that satisfies, 
\begin{eqnarray}
\int_{-\infty}^{f(x)}I(\tau)d\tau&=&\int_{-\infty}^{x}I_0(\tau)d\tau,
\end{eqnarray}
which implies that $f'(I\circ f)=I_0$, where $f'=\frac{\partial f}{\partial x}$. For continuous and positive probability densities $I_0$ and $I$, $f$ is a strictly increasing function and is defined uniquely. Note that $f$ morphs the input signal $I$ into the reference signal $I_0$, through $f'(I\circ f)=I_0$.

 In \cite{serim2015} we showed that the CDT can enhance linear separability of signal classes in transform (i.e. feature) space. A simple example to demonstrate the linear separability characteristic of the CDT is as follows. Let $I:\R\rightarrow \R^+$ be a signal and let $\It$ be its corresponding representation in the CDT space with respect to a chosen template signal $I_0:\R\rightarrow\R^+$. If we consider the CDT transform of the translated signal (see \cite{serim2015} for a derivation), we have that
  \begin{eqnarray}
  I(t-\tau)\operatorname*{\longrightarrow}^{CDT} \It(t)+\tau\sqrt{I_0(t)}, ~\forall t,\tau\in\R.
\label{eq:cdttrans}
 \end{eqnarray} 
Now observe that although $I(t-\tau)$ is nonlinear in $\tau$, its CDT representation $\It(t)+\tau\sqrt{I_0(t)}$ becomes linear in $\tau$. This effect is not limited to translations and is generalized to larger classes of signal transformations. More precisely, let $\C$ be a set of measurable maps and let $\P$ and $\Q$ be sets of positive probability density functions born from two positive probability density functions $p_0$ and $q_0$ (mother density functions) as follows,
\begin{eqnarray}
\P&=&\{p| p=h'(p_0\circ h), \forall h\in\C \},\nonumber\\
\Q&=&\{q| q=h'(q_0\circ h), \forall h\in\C \}.
\end{eqnarray}
The sets $\P$ and $\Q$ are linearly separable in the transform space (regardless of the choice of the reference signal $I_0$) if $\C$ satisfies the following conditions, 
\begin{enumerate}[i)]
\item $h\in\C \iff h^{-1}\in \C$
\item $h_1, h_2\in \C \Rightarrow \alpha h_1+(1-\alpha) h_2 \in \C, ~\forall\alpha\in[0,1]$
\item $h_1, h_2\in \C \Rightarrow h_1(h_2), h_2(h_1)\in\C$ 
\item $h'(p_0\circ h)\neq q_0,~\forall h\in\C$ 
\end{enumerate}

Finally, we note that the CDT has well-understood geometric properties. The Euclidean norm of the signal $I$ in the transform space corresponds to the 2-Wasserstein distance between $I$ and $I_0$, which is given by
\begin{eqnarray}
\|\It\|_2= (\int_\R (f(x)-x)^2I_0(x)dx)^\frac{1}{2},
\end{eqnarray} 
where $f' (I\circ f)=I_0$. Note that, using the optimal transportation (OT) parlance, in one-dimensional problems there only exist  one strictly increasing transport map $f$ that morphs $I$ into $I_0$ \cite{villani2008optimal} and hence no optimization, over $f$, is required to calculate the 2-Wasserstein distance. 

\subsection{The Radon transform}

\begin{figure}[t]
\centering
\includegraphics[width=.85\columnwidth]{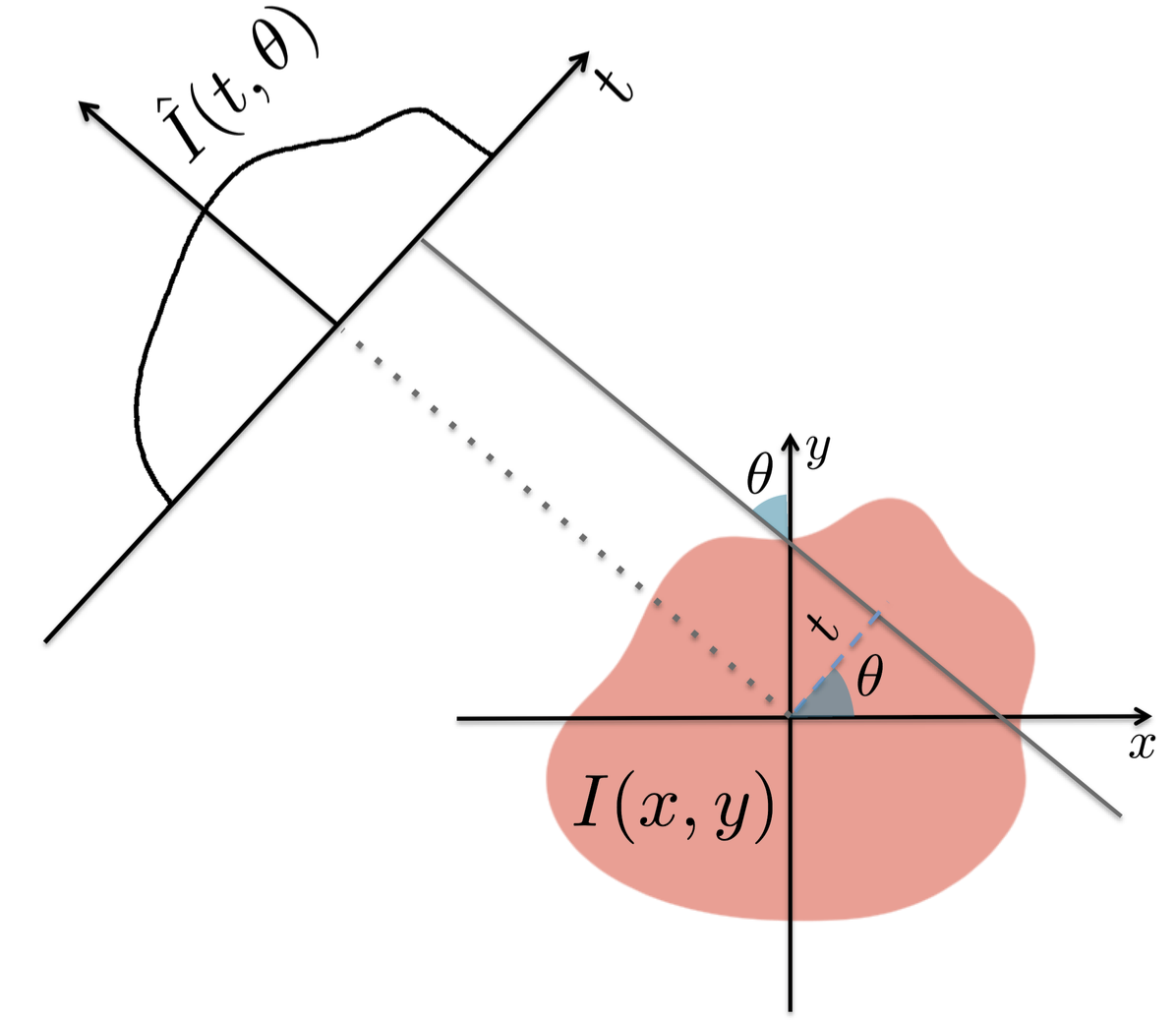}
\caption{Geometry of the line integral associated with the Radon transform}
\label{fig:radon}
\end{figure}

The Radon transform of an image $I:\R^2\rightarrow\R^{+}$, which we denote by $\Ih=\Rad(I)$, is defined as: 
\begin{eqnarray}
\Ih(t,\theta)&=&\int_{-\infty}^{\infty}\int_{-\infty}^{\infty} I(x,y)\delta(t-x\cos(\theta)-y\sin(\theta))dxdy\nonumber\\
\end{eqnarray}
where $t$ is the perpendicular distance of a line from the origin and $\theta$ is the angle between the line and the y-axis as shown in Figure \ref{fig:radon}. Furthermore, using the Fourier Slice Theorem \cite{quinto2006introduction,natterer1986mathematics}, the inverse Radon transform is defined as, $I=\iRad(\Ih)$,
\begin{eqnarray}
I(x,y)&=&\int_0^\pi (\Ih(.,\theta)*w(.))\circ(x\cos(\theta)+y\sin(\theta))d\theta\nonumber\\
\end{eqnarray}
where $w=\F^{-1}(|\omega|)$ is the ramp filter, $\F^{-1}$ is the inverse Fourier transform, and $\Ih(.,\theta)*w(.)$ is the one-dimensional convolution with respect to variable $t$. We will use the following property of the Radon transform in our derivations in the consequent sections, 
\begin{eqnarray}
\int_{-\infty}^{\infty}\int_{-\infty}^{\infty} I(x,y)dxdy =\int_{-\infty}^{\infty}\Ih(t,\theta)dt, \;\;\;\;\; \forall\theta\in[0,\pi]
\label{eq:Rtheta}
\end{eqnarray}
which implies that $\int_{-\infty}^{\infty}\Ih(t,\theta_i)dt=\int_{-\infty}^{\infty}\Ih(t,\theta_j)dt$ for $\forall\theta_i,\theta_j\in[0,\pi]$.

\section{The Radon-CDT}
\label{sec:Radon-CDT}

Here we combine the CDT \cite{serim2015} and the Radon transform to describe the Radon Cumulative Distribution Transform (Radon-CDT). We then derive a few properties of the Radon-CDT, and extend the CDT results \cite{serim2015} on linear separability of classes of one-dimensional signals  \cite{serim2015} to classes of images. Before introducing Radon-CDT we  first introduce a metric for images, which we call the Radon Cumulative Distribution (RCD) metric. 


 Let $\mu$ and $\sigma$  be two continuous probability measures on $\R^2$ with corresponding positive probability density functions $I$ and $I_0$.  Let the sinograms obtained from their respective Radon transforms be, 
 \begin{eqnarray}
 \left\{ \begin{array}{l}
 \Ih_0=\Rad(I_0)\\
 \Ih=\Rad(I)
 \end{array}\right.
 \end{eqnarray}
Using the Radon property shown in Eq.\eqref{eq:Rtheta}, for a fixed angle $\theta$, there exists a unique one-dimensional measure preserving map, $f(.,\theta)$ that warps $\Ih(.,\theta)$ into $\Ih_0(.,\theta)$ and satisfies the following:
\begin{eqnarray}
\int_{-\infty}^{f(t,\theta)}\Ih(\tau,\theta)d\tau=\int_{-\infty}^{t}\Ih_0(\tau,\theta)d\tau,~ \forall \theta\in[0,\pi]
\label{eq:fi}
\end{eqnarray}
which implies that $f'(.,\theta)(\Ih(.,\theta)\circ f(.,\theta))=\Ih_0(.,\theta)$. Using $f$ we define the RCD metric between images $I$ and $I_0$ as,
\begin{eqnarray}
d_{RCD}(I,I_0)= \left(\int_0^\pi \int_{-\infty}^{\infty} (f(t,\theta)-t)^2\Ih_0(t,\theta) dtd\theta\right) ^{\frac{1}{2}}.
\end{eqnarray}
In Appendix \ref{sec:appendix} we show that $d_{RCD}:\R^2\times\R^2\rightarrow \R^+_0$ satisfies the non-negativity, coincidence axiom, symmetry, and triangle inequality properties and therefore is a metric. We note that the RCD metric as defined above is also known as the sliced Wasserstein metric in the literature and is used in \cite{rabin2012wasserstein,bonneel2015sliced} to calculate barycenters of measures for texture mixing applications.

We now define the Radon-CDT. Given an image $I$ and a template image $I_0$, where both images are normalized such that
\begin{eqnarray*}
\int_{\R^2} I(\x)d\x =\int_{\R^2} I_0(\x)d\x =1,
\end{eqnarray*}
the forward and inverse Radon-CDT for image $I$ are defined as, 
\begin{eqnarray}
\left\{
\begin{array}{l}
\It(.,\theta)=(f(.,\theta)-id)\sqrt{\Ih_0(.,\theta)}\\
\\
I= \Rad^{-1}( det(D\g)( \Ih_0\circ \g))
\end{array}
\right.
\label{eq:Radon-CDT}
\end{eqnarray}
 where $\g(t,\theta)=[f^{-1}(t,\theta),\theta]^T$, and $D\g$ is the Jacobian of $\g$. In order to avoid any confusion, we emphasize that $f^{-1}(f(.,\theta),\theta)=id,~\forall \theta\in[0,\pi]$, and $det(D\g (t,\theta))=\frac{\partial f^{-1}(t,\theta)}{\partial t}$.
 
\begin{figure}
 \centering
 \includegraphics[width=\columnwidth]{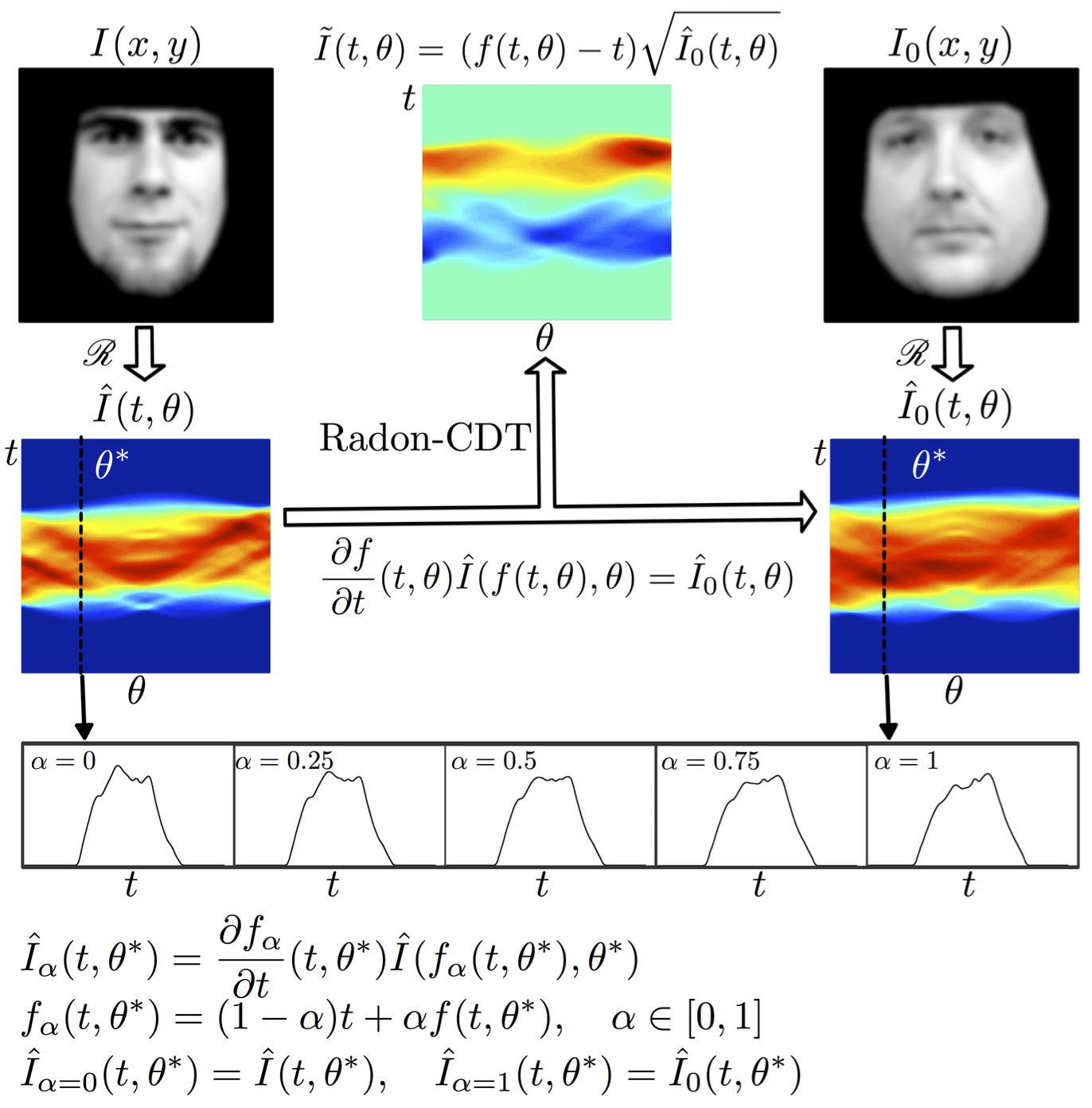}
 \caption{The process of calculating the Radon-CDT transform of image $I$ with respect to the template image $I_0$.}
 \label{fig:Radon-CDT}
 \end{figure}

  Figure \ref{fig:Radon-CDT} shows the process of calculating the Radon-CDT of a sample image $I$ with respect to a template image $I_0$. The sinograms of images are first computed and denoted as $\Ih$ and $\Ih_0$. Then for each $\theta$ the measure preserving map, $f(.,\theta)$, is found to warp the one-dimensional signal $\Ih(.,\theta)$ into $\Ih_0(.,\theta)$. The one-dimensional warping between $\Ih(.,\theta^*)$ and $\Ih_0(.,\theta^*)$, where $\theta^*$ is an arbitrary projection angle, is visualized in Figure \ref{fig:Radon-CDT} to demonstrate the process. Finally, the Radon-CDT is obtained from $f$ and $\Ih_0$.
 
Similar to the CDT, the Radon-CDT is a nonlinear isomorphic image transform, since for a given template image $I_0$, $f$ in Eq. \eqref{eq:fi} provides a unique representation of $I$.  Furthermore, the Euclidean norm of image $I$ in the transformed space corresponds to the RCD metric between the image and the reference image,
 \begin{eqnarray}
 \| \It \|_2&=& \left(\int_0^\pi \int_{-\infty}^{\infty} (f(t,\theta)-t)^2\Ih_0(t,\theta) dtd\theta\right)^{\frac{1}{2}} \nonumber\\
 &=& d_{RCD}(I,I_0).
 \end{eqnarray}
In addition, the Euclidean distance between two images $I_i$ and $I_j$ in the transformed space is also the RCD metric between these images, 
{\small
 \begin{eqnarray}
 \| \It_i-\It_j \|_2&=& \left(\int_0^\pi \int_{-\infty}^{\infty} (f_i(t,\theta)-f_j(t,\theta))^2\Ih_0(t,\theta) dtd\theta\right)^{\frac{1}{2}} \nonumber\\
 &=& d_{RCD}(I_i,I_j).
 \end{eqnarray}}
The proof for the equation above is included as part of the proof for the triangle inequality property of the Radon-CDT metric in Appendix \ref{sec:appendix}. 

 \subsection{Radon-CDT properties}
 Here we describe a few basic properties of the Radon-CDT, with the main purpose of elucidating certain of its qualities necessary for understanding its ability to linearly separate certain types of two-dimensional densities.
 
{\it\bf Translation.} Let $J(x,y)=I(x-x_0,y-y_0)$ and let $\It$ be the Radon-CDT of $I$. The Radon-CDT of $J$ with respect to a reference image $I_0$ is given by,
\begin{eqnarray}
\tilde{J}(t,\theta)=\It(t,\theta)+ (x_0\cos(\theta)+y_0\sin(\theta))\sqrt{\Ih_0(t,\theta)},\nonumber\\ ~~t\in\R~\text{and}~\theta\in[0,\pi].
\label{eq:transCDT}
\end{eqnarray}
For a proof, see Appendix \ref{sec:ptrans}. Similar to the CDT example in Eq. \ref{eq:cdttrans}, it can be seen that while $I(x-x_0,y-y_0)$ is nonlinear with respect to $[x_0,y_0]$  the presentation of the image in the Radon-CDT, $\It(t,\theta)+ (x_0\cos(\theta)+y_0\sin(\theta))\sqrt{\Ih_0(t,\theta)}$ is linear.

{\it \bf Scaling.} Let $J(x,y)=\alpha^2I(\alpha x,\alpha y)$ with $\alpha>0$ and let $\It$ be the Radon-CDT of $I$. The Radon-CDT of $J$ with respect to a reference image $I_0$ is given by,
\begin{eqnarray}
\tilde{J}(t,\theta)=\frac{\It(t,\theta)}{\alpha}+ (\frac{1-\alpha}{\alpha})\sqrt{\Ih_0(t,\theta)},\nonumber\\ ~~t\in\R~\text{and}~\theta\in[0,\pi].
\end{eqnarray}
For a proof, see Appendix \ref{sec:pscale}. Similar to the translation property, it can be seen that while $\alpha^2I(\alpha x,\alpha y)$ is nonlinear with respect to $\alpha$ the corresponding presentation in the Radon-CDT space, $\frac{\It(t,\theta)}{\alpha}+ (\frac{1-\alpha}{\alpha})\sqrt{\Ih_0(t,\theta)}$,  is linear in $\frac{1}{\alpha}$.

{\it \bf Rotation.} Let $J(x,y)=I(x\cos(\phi)+y\sin(\phi),-x\sin(\phi)+y\cos(\theta))$ and let $\It$ be the Radon-CDT of $I$. For a circularly symmetric reference image $I_0$, the Radon-CDT of $J$ is given by,
\begin{eqnarray}
\tilde{J}(t,\theta)=\It(t,\theta-\phi), ~~t\in\R~\text{and}~\theta\in[0,\pi]
\end{eqnarray}
for a proof, see Appendix \ref{sec:prot}. Note that unlike translation and scaling, for rotation the transformed image remains nonlinear with respect to $\phi$.
 
  \subsection{Linear separability in the Radon-CDT space}
  \label{sec:lRadon-CDT}
 In this section we describe how the Radon-CDT can enhance linear separability of image classes.  The idea is to show that if a specific `generative model' is utilized for constructing signal classes, the Radon-CDT can be effective in linearly classifying these.  Let $\C$ be a set of measurable maps, with $\h \in \C$, and let $\P$ and $\Q$ be sets of normalized images born from two mother images $p_0$ and $q_0$ as follows,
 \begin{eqnarray}
\P&=&\{p| p=\iRad(det(D\h)(\hat{p}_0\circ \h)), \nonumber\\
&&~~~ \h(t,\theta)=[h(t,\theta),\theta]^T, \forall h\in\C, \forall\theta\in[0,\pi] \},\nonumber\\
\Q&=&\{q| q=\iRad(det(D\h)(\hat{q}_0\circ \h)), \nonumber\\
&&~~~ \h(t,\theta)=[h(t,\theta),\theta]^T, \forall h\in\C, \forall\theta\in[0,\pi] \}.
\label{eq:PQ}
\end{eqnarray}
Before proceeding, it is important to note that $h$ must be absolutely continuous in $t$ and $\theta$, so that $det(D\h)(\hat{p}_0\circ \h)$ and  $det(D\h)(\hat{q}_0\circ \h)$ remain in the range of the Radon transform \cite{gelfand}. Now, under the signal generative model described above, it can be shown that the sets $\P$ and $\Q$ become linearly separable in the transform space (regardless of the choice of the reference image $I_0$) if $\C$ satisfies the following conditions, 
\begin{enumerate}[i)]
\item $h\in\C \iff h^{-1}\in \C$
\item $h_1, h_2\in \C \Rightarrow \alpha h_1+(1-\alpha) h_2 \in \C, ~\forall\alpha\in[0,1]$
\item $h_1, h_2\in \C \Rightarrow h_1(h_2), h_2(h_1)\in\C$ 
\item $det(D\h)(\hat{p}_0 \circ \h)\neq \hat{q}_0,~\h(t,\theta)=[h_\theta(t),\theta]^T,~\forall h_\theta\in\C$ 
\end{enumerate}
The proof is included in Appendix \ref{sec:separ}. 

It is useful to consider a couple of examples to elucidate the meaning of the result above. Consider for example $\C=\{h| h(t,\theta)=t+x_0\cos(\theta)+y_0\sin(\theta), \forall x_0,y_0\in \R \}$ which corresponds to the class of translations in the image space (i.e. translation by $[x_0,y_0]$).  Such a class of diffeomorphisms satisfies all the conditions named above, and if applied to two mother signals $\hat{p}_0$ and $\hat{q}_0$ would generate signals in image space which are simple translations of $p_0$ and $q_0$. The classes $\P$ and $\Q$ would therefore not be linearly separable in signal domain. The result above, however, states that these classes are linearly separable in Radon-CDT domain. Another example is the class $\C=\{h| h(t,\theta)=\beta t, \forall \beta\in\R^+\}$ which corresponds to the class of mass preserving scalings in the image space. It is straightforward to show that such class of mass preserving mappings also satisfies the conditions enumerated above.  These cases only serve as few examples that represent such classes of mass preserving mappings.  An important aspect to the theory presented above is that the linear separation result is independent of the choice of template $I_0$ used in the definition of the transform. Therefore, in theory, linear separability could be achieved utilizing any chosen template.

\section{Numerical implementation}
\label{sec:numerics}
\subsection{Radon transform:}

A large body of work on numerical implementation of the Radon transform exists in the literature \cite{averbuch2001fast}. Here, we use a simple numerical integration approach that utilizes nearest neighbor interpolation of the given images, and summation.  In all our experiments we used 180 projections (i.e. $\theta_i= (i-1)^\circ$, $i\in[1,...,180]$). 

\subsection{Measure preserving map:}
In this section, we follow a similar computational algorithm as in Park et al. \cite{serim2015} and describe a numerical method to estimate the measure preserving map that warps $\Ih(.,\theta)$ into $\Ih_0(.,\theta)$. Let $\pi$ be the B-spline of degree zero of width $r$,
\begin{eqnarray}
\pi(x)=\left\{\begin{array}{lc}
\frac{1}{r} & |x|\leq \frac{1}{2}r\\
0 & |x|>\frac{1}{2}r
 \end{array}\right.
\end{eqnarray}
and define $\Pi$ as,
\begin{eqnarray}
\Pi(x)=\left\{\begin{array}{lc}
0 & x<-\frac{1}{2}r\\
\frac{x}{r}+\frac{1}{2} & -\frac{1}{2}r\leq x \leq\frac{1}{2}r\\
1& x>\frac{1}{2}r
 \end{array}\right.
\end{eqnarray}
Using the B-spline of degree zero, we approximate the continuous sinograms, $\Ih(.,\theta)$ and $\Ih_0(.,\theta)$, with their corresponding discrete counterparts $\c$ and $\c_0$ as follows,
\begin{eqnarray}
\left\{\begin{array}{l}
\Ih(t,\theta)\approx\sum_{k=1}^K \c[k]\pi(t-t_k)\\
\Ih_0(t,\theta)\approx\sum_{k=1}^K \c_0[k]\pi(t-t_k)
\end{array}\right.
\end{eqnarray}
Now the goal is to find $f(.,\theta)$ such that, 
\begin{eqnarray}
\int_{-\infty}^{f(t,\theta)} \Ih(\tau,\theta)d\tau=\int_{-\infty}^{t} \Ih_0(\tau,\theta)d\tau
\end{eqnarray}
which is equivalent to,
\begin{eqnarray}
\sum_{k=1}^K \c[k]\Pi(f(t,\theta)-t_k)=\sum_{k=1}^K \c_0[k]\Pi(t-t_k)
\end{eqnarray}
let $\brho=[0,\frac{1}{L},\frac{2}{L},...,\frac{L-1}{L},1]^T$ for $L>1$, and define $\btau_0$ and $\btau$ such that,   
\begin{eqnarray}
\left\{\begin{array}{l}
\sum_{k=1}^K \c[k]\Pi(\btau[l]-t_k)=\brho[l]\\
\sum_{k=1}^K \c_0[k]\Pi(\btau_0[l]-t_k)=\brho[l]
\end{array}\right. 
\end{eqnarray}
for $l=1,...,L+1$, where $\btau$ and $\btau_0$ are found using the algorithm defined in Park et al. \cite{serim2015}. From the equation above we have that $f(\btau_0[l],\theta)=\btau[l]$. Finally we interpolate $f$ to obtain its values on the regular grid, $t_k$ for $k=1,...,K$. 
\subsection{ Computational complexity}
 The computational complexity of the Radon transform of $N\times N$ images at $M$ projection angles is $\O(N^2M)$, and the computational cost for finding the mass preserving map,  $f(t,\theta)$, from a pair of sinograms is $\O(MN log(N))$, hence, the overall computational cost of the Radon-CDT is dominated by the computational complexity of the Radon transform, $\O(N^2M)$. We also compare our image transform with the Ridgelet transform. The Ridgelet transform can be presented as the composition of the Wavelet transform and the Radon transform. Since the computational complexity of the Wavelet transform is $\O(N^2)$, the computational complexity of the Ridgelet transform is also dominated by the computational complexity of the Radon transform, $\O(N^2M)$.

\begin{figure}
\centering
\includegraphics[width=\columnwidth]{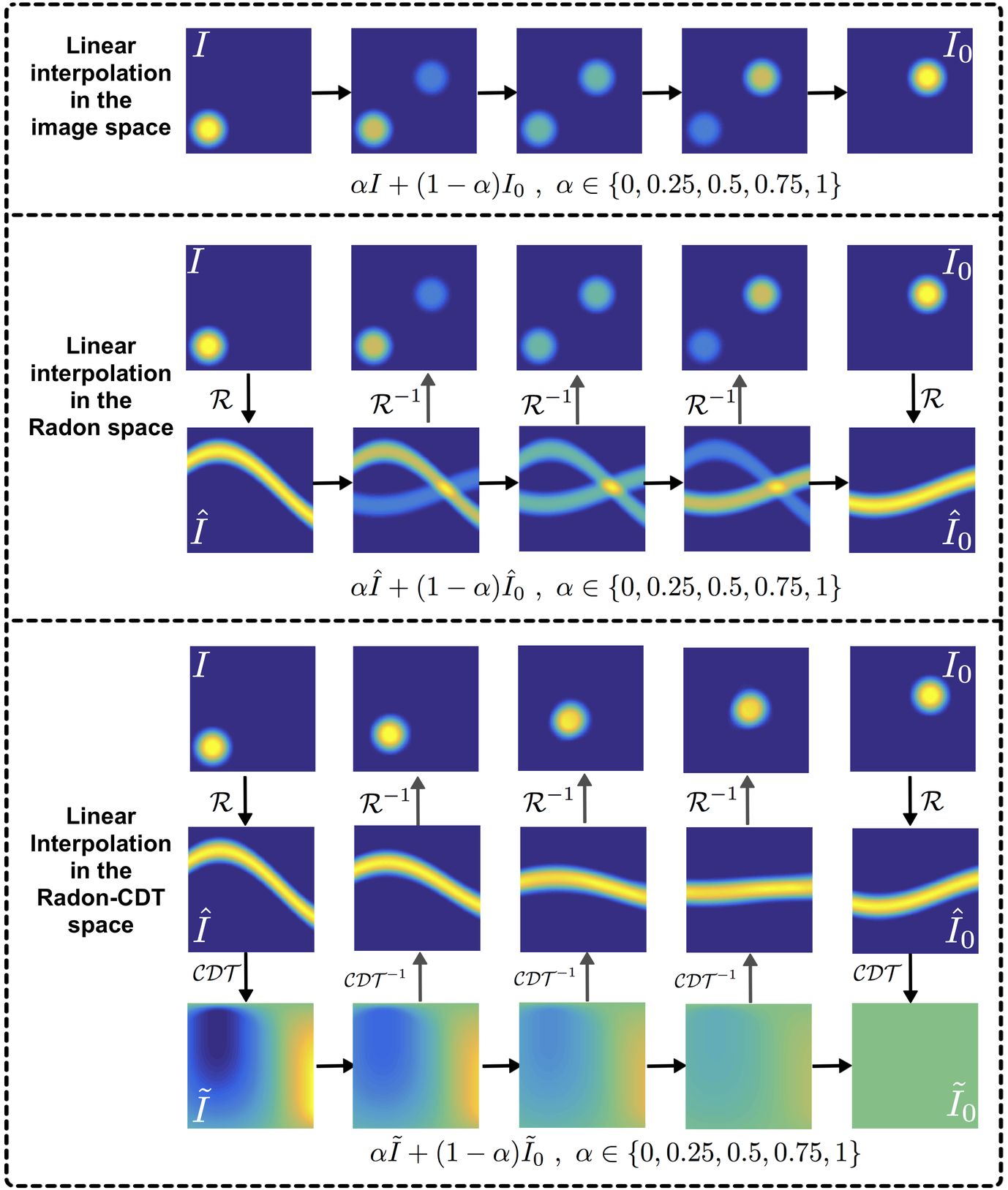}
\caption{A simple linear interpolation between two images in the image space, the Radon transform space (which is a linear transform), and the Radon-CDT space.}
\label{fig:interpolation}
\end{figure}

\section{Results}
\label{sec:results}

 In this section, we start by demonstrating the invertible and nonlinear nature of the Radon-CDT. We first show that Radon-CDT provides a strong framework for modeling images. Then, we study the ability of the Radon-CDT to enhance linear separability in a variety of pattern recognition tasks. Starting with a simple synthetic example,  we explain the idea of linear separation in the Radon-CDT space. Next, we describe the application of Radon-CDT in pattern recognition by demonstrating its capability to simplify data structure on three image datasets.  The first dataset is part of the Carnegie Mellon University Face Images database, as described in detail in \cite{stegmann2003fame}, and it includes frontal images of 40 subjects under neutral and smiling facial expressions. The second dataset contains 500 images of segmented liver nuclei extracted from histology images obtained from the archives of the University of Pittsburgh Medical Center (UPMC). The nuclei belong to 10 different subjects including five cancer patients suffering from fetal-type hepatoblastoma (FHB), with the remaining images from the liver of five healthy individuals \cite{wang2010detection} (in average 50 nuclei are extracted per subject). The last dataset is part of the LHI dataset of animal faces \cite{si2012learning}, which includes $159$ images of cat faces, $101$ images of deer faces, and $116$ images of panda faces. 

\subsection{The Radon-CDT representation}
 In the previous sections we demonstrated few properties of the Radon-CDT. Here we show some implications of these properties in image modeling. Let $I_0$ be an arbitrary image and let $I(x,y)=I_0(x-x_0,y-y_0)$ be a translated version of $I_0$.  A natural interpolation between these images follows from $I_\alpha(x,y)=I_0(x-\alpha x_0, y-\alpha y_0)$ where $\alpha\in[0,1]$. The linear interpolation between these images in the image space, however, is equal to, 
\begin{eqnarray}
I_\alpha(x,y)&=&\alpha I_1(x,y)+(1-\alpha)I_0(x,y)\nonumber\\
&\neq& I_0(x-\alpha x_0, y-\alpha y_0)			
\end{eqnarray} 
In fact, above equation is also true for any linear image transform. Take the Radon transform for example, where the linear interpolation in the transform space is equal to,
\begin{eqnarray}
\hat{I}_\alpha(t,\theta)&=&\alpha \Rad(I_1(x,y))+(1-\alpha)\Rad(I_0(x,y))\nonumber\\
&=& \Rad( \alpha I_1(x,y))+(1-\alpha)I_0(x,y)) \Rightarrow \nonumber \\
I_\alpha(x,y)&=& \iRad(\hat{I}_\alpha(t,\theta))= \alpha I_1(x,y)+(1-\alpha)I_0(x,y) \nonumber\\
&\neq& I_0(x-\alpha x_0, y-\alpha y_0).
\label{eq:linearInterp}
\end{eqnarray}
On the other hand, due to its nonlinear nature, this scenario is completely different in the Radon-CDT space. Let $I_0$ be the template image for the Radon-CDT (the following argument holds even if the template is chosen to be different from $I_0$), then from Equation \eqref{eq:transCDT} we have $\tilde{I}_0(t,\theta)=0$ and  $\tilde{I}_1(t,\theta)=(x_0cos(\theta)+y_0sin(\theta))\sqrt{\hat{I}_0(t,\theta)}$. The linear interpolation in the Radon-CDT space is then equal to,
\begin{eqnarray}
\tilde{I}_\alpha(t,\theta)&=&\alpha \tilde{I}_1(t,\theta)+(1-\alpha)\tilde{I}_0(t,\theta)\nonumber\\
&=& \alpha (x_0 cos(\theta)+ y_0 sin(\theta))\sqrt{\hat{I}_0(t,\theta)} \Rightarrow \nonumber\\
I_\alpha(t,\theta)&=& I_0(x-\alpha x_0, y-\alpha y_0).
\end{eqnarray} 
which is the natural interpolation between these images and captures the underlaying translation. Figure \ref{fig:interpolation} summarizes the equations presented above  and provides a visualization of this effect.  

\begin{figure}[h]
\centering
\includegraphics[width=\columnwidth]{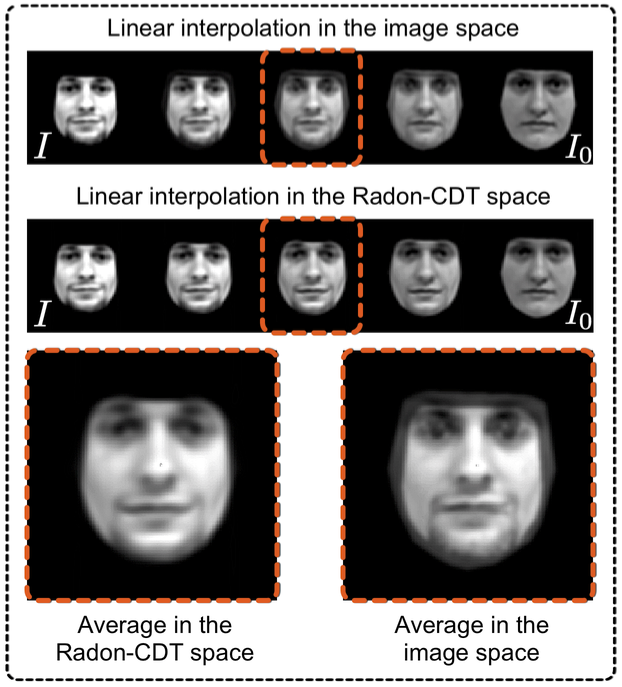}
\caption{Interpolation in the image space (or any linear transform space) and in the Radon-CDT space. The corresponding average images in these spaces demonstrate the benefit of modeling images through our proposed nonlinear and invertible transform.}
\label{fig:FaceInterpolation}
\end{figure}

In fact, translation and scaling are not the only effects that are captured by our proposed transform. The Radon-CDT is capable of capturing more complicated variations in the image datasets. In order to demonstrate the modeling (or representation) power of the Radon-CDT we repeat the experiment above on two face images taken from the Carnegie Mellon University Face Images database. Figure \ref{fig:FaceInterpolation} shows the interpolated faces in the image space and in the Radon-CDT space. From Figure \ref{fig:FaceInterpolation} it can be seen that the nonlinearity of the Radon-CDT enables it to capture variations in a much more efficient way (this will also be demonstrated in the subsequent sections). Note that according to Equation \eqref{eq:linearInterp} the interpolation in any linear transform space (such as the Radon transform or the Ridgelet transform) leads to the same interpolation in the image space as shown in Figure \ref{fig:FaceInterpolation}.

\begin{figure}[h]
\centering
\includegraphics[width=\columnwidth]{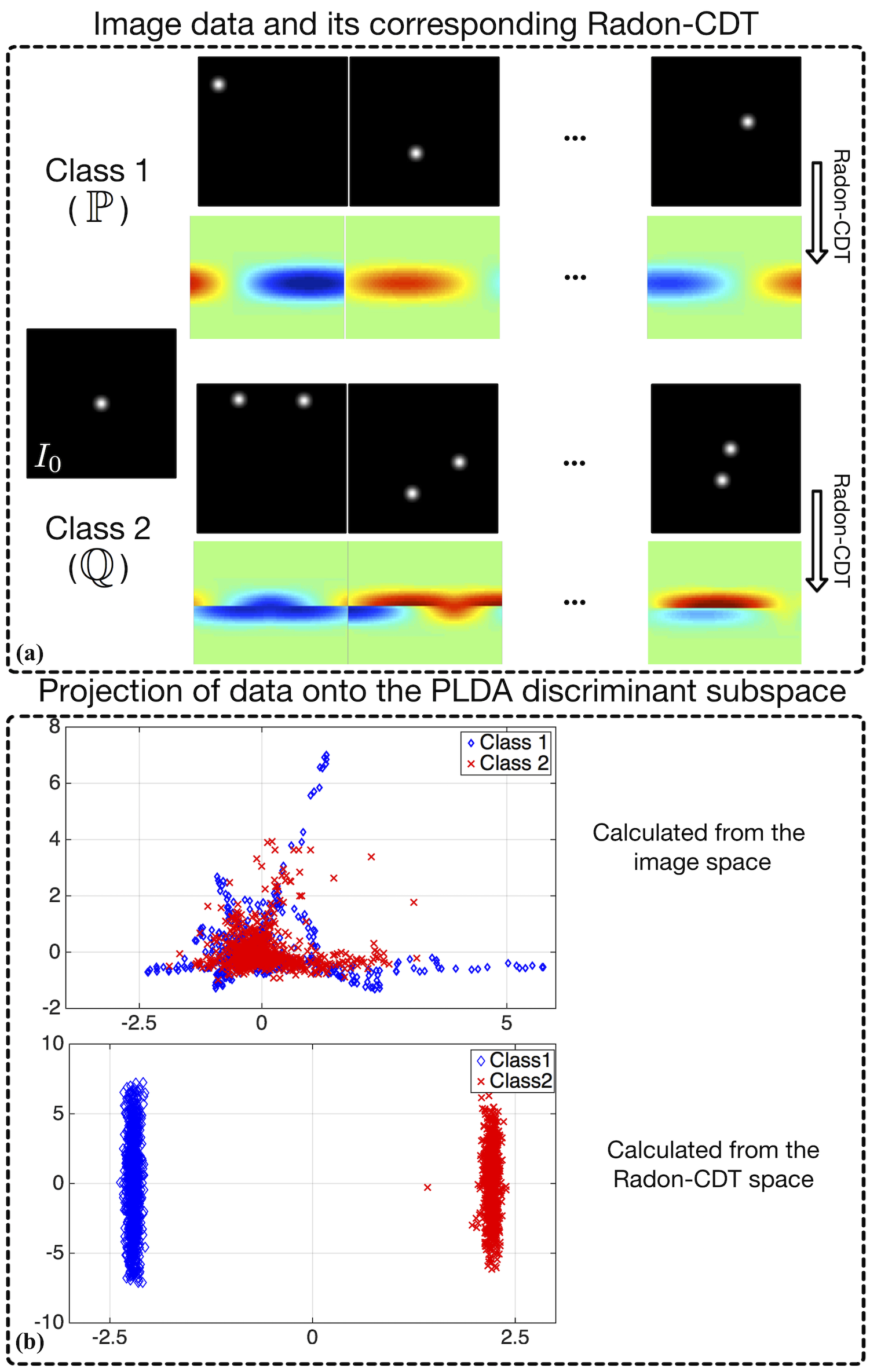}
\caption{Two example image classes $\P$ and $\Q$ and their corresponding Radon-CDT with respect to the template image $I_0$ (a), and the projection of the data and its transformation onto the {\it p}LDA discriminant subspace learned from the image space (top) and the Radon-CDT space (bottom), respectively (b). }
\label{fig:res1}
\end{figure}

\subsection{Synthetic example}
\label{sec:synthetic}
Consider two classes of images $\P$ and $\Q$ which are generated as follows,
\begin{eqnarray}
\P&=&\{p| p(\x)=\frac{1}{2\pi\sigma^2} e^{-\frac{\|\x-\bmu\|^2}{2\sigma^2}},\bmu\sim\text{unif}([0,1]^2)\} \nonumber\\
\Q&=&\{q| q(\x)=\frac{1}{4\pi\sigma^2} (e^{-\frac{\|\x-\bmu_1\|^2}{2\sigma^2}}+e^{-\frac{\|\x-\bmu_2\|^2}{2\sigma^2}}),\nonumber\\
&&~~~~~~~~~~~~~~\bmu_1,\bmu_2\sim\text{unif}([0,1]^2), \bmu_1\neq\bmu_2\}
\end{eqnarray}
Figure \ref{fig:res1}(a) illustrates these classes of images. Classes $\P$ and $\Q$ are disjoint, however, they are not linearly separable in the image space. This is demonstrated by projecting the image classes onto a linear discriminant subspace, calculated using penalized linear discriminant analysis ({\it p}LDA) \cite{wang2011penalized}, which is a regularized version of LDA. More precisely, we first prune the image space by discarding dimensions which do not contain data points. This is done using principle component analysis (PCA) and discarding the zero eigenvalues. Then, we calculate the {\it p}LDA subspace from the pruned image space. Figure \ref{fig:res1}(b) shows the projection of the data onto the subspace spanned by the first two {\it p}LDA directions. 

Next, we demonstrate the linear separability property of our proposed image transform  by  calculating the Radon-CDT of classes $\P$ and $\Q$ with respect to an arbitrary image $I_0$ (see Figure \ref{fig:res1}(a)), and finding the {\it p}LDA subspace in the transformed space. Projection of the transformed data, $\tilde{\P}$ and $\tilde{\Q}$, onto the {\it p}LDA subspace (as depicted in Figure \ref{fig:res1}(b)) indicates that the nonlinearity of the data is captured by Radon-CDT and the image classes have become linearly separable in the Radon-CDT space.  

\subsection{Pattern recognition in the Radon-CDT space}
In this section we investigate the linear separability property of Radon-CDT on real images, where the image classes do not exactly follow the class structures stated in Section \ref{sec:lRadon-CDT}. As with the simulated example above, our goal is to demonstrate that the data classes in the Radon-CDT space become more linearly separable. This is done by utilizing linear support vector machine (SVM) classifiers in the image and the Radon-CDT space, and showing that the linear classifiers consistently lead to higher classification accuracy in the Radon-CDT space.  

\begin{figure*}[t]
\centering
\includegraphics[width=\linewidth]{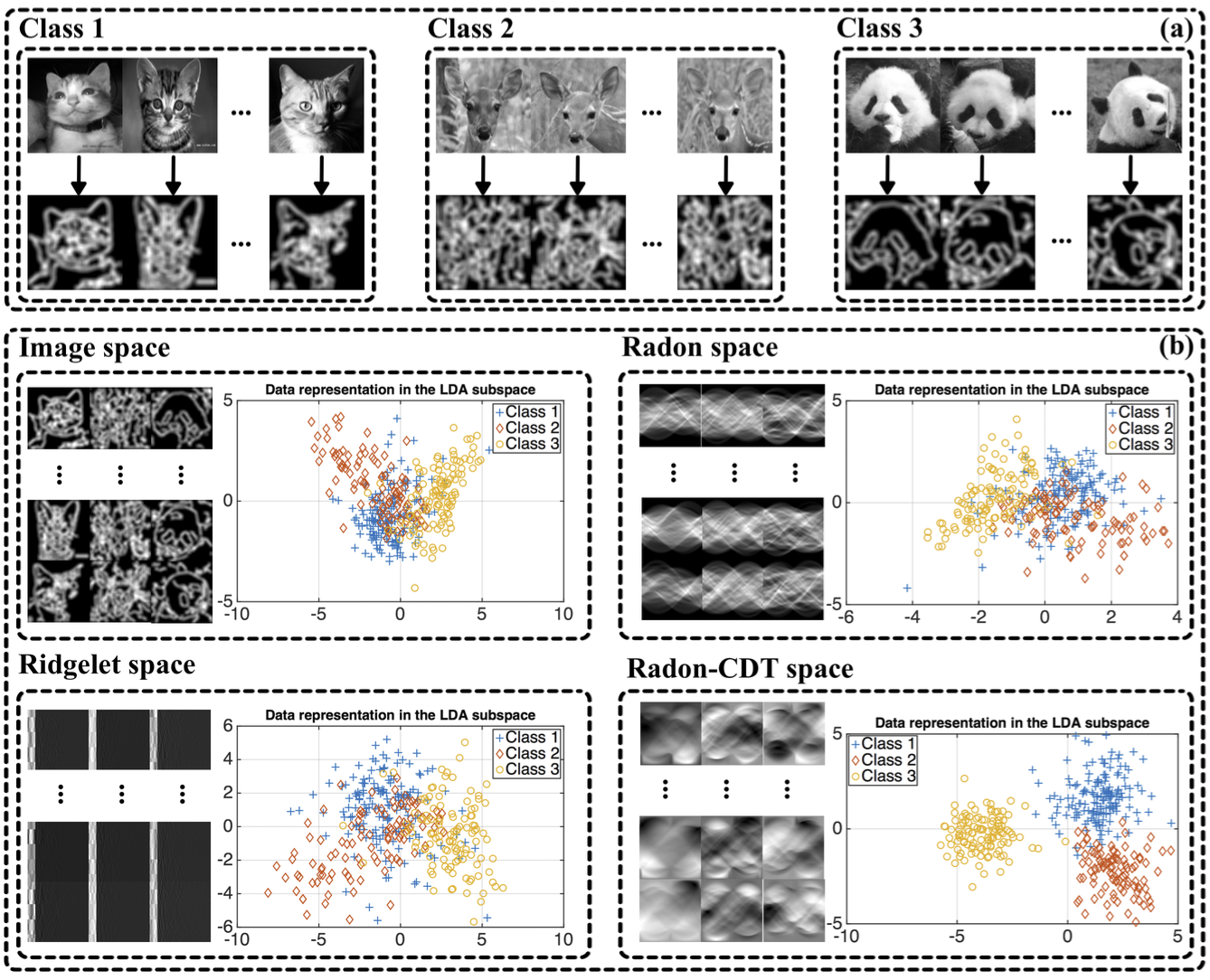}
\caption{ Sample images belonging to each class and their corresponding edge-maps (a), and the projection of the data and its transformations onto the {\it p}LDA discriminant subspace learned from the image space, the Radon transform space, the Ridgelet transform space, and the Radon-CDT space (b).}
\label{fig:res01}
\end{figure*}

\begin{figure*}[t]
\centering
\includegraphics[width=\linewidth]{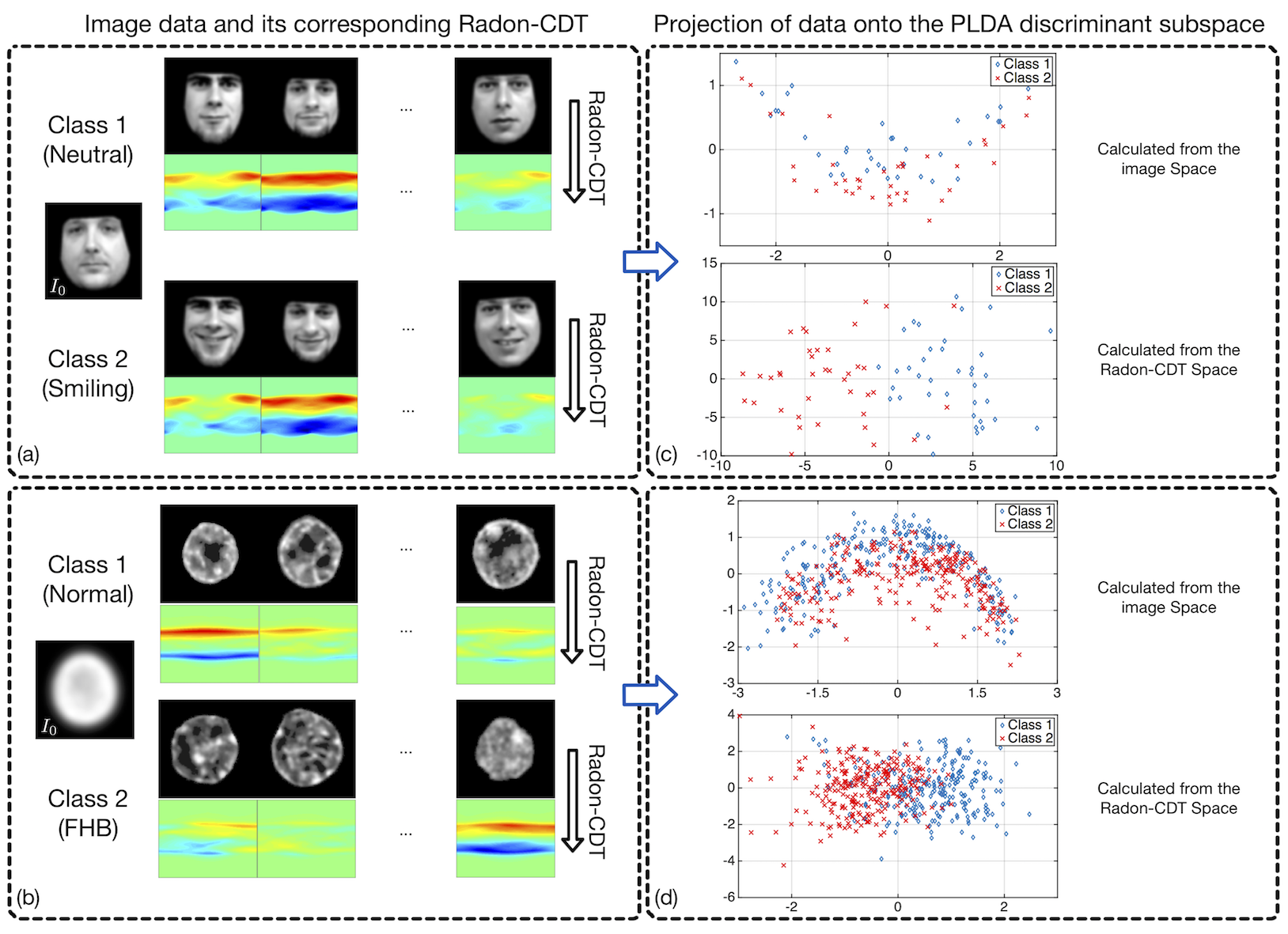}
\caption{The image classes and their corresponding Radon-CDT with respect to the template image $I_0$ for the facial expression (a) and the liver nuclei (b) datasets, and the projection of the data and its transformation onto the {\it p}LDA discriminant subspace learned from the image space and the Radon-CDT space, for the facial expression (c) and the liver nuclei (d) dataset.}
\label{fig:res2}
\end{figure*}

To test our method, we utilized a facial expression dataset \cite{stegmann2003fame}, a liver nuclei dataset \cite{basu2014detecting}, and part of the LHI dataset of animal faces \cite{si2012learning}. The facial expression dataset contains two classes of expressions, namely `neutral' and `smiling'. The classes in the nuclei dataset are `fetal-type hepatoblastoma' (type of a liver cancer) and `benign' for liver nuclei. The last dataset contains facial images of three different animals, namely cat, deer, and panda under a variety of variations including translation, pose, scale, texture, etc. The animal face dataset is preprocessed by first calculating the image edges using the Canny operator and then filtering the edge-maps of the images with a Gaussian low pass filter.

 We compare our Radon-CDT with well-known image transforms such as the Radon transform and the Ridgelet transform \cite{do2003finite}. Figure \ref{fig:res01} shows sample images from the LHI dataset, the corresponding Radon transform, the Ridgelet transform, and the Radon-CDT of the images. The Radon-CDT  and the Ridgelet transform are calculated at discrete projection angles, $\theta\in [0^\circ,1^\circ,...,179^\circ]$, and $3$ levels were used for the Ridgelet transform. In addition, Figure \ref{fig:res01} depicts the discriminant subspaces calculated for this dataset in all transformation spaces. The discriminant subspaces are calculated  using the {\it p}LDA, as described in Section \ref{sec:synthetic}. It can be clearly seen that the image classes become more linearly separable in the Radon-CDT space.

 Similarly,  sample images from the facial expression dataset, the nuclei dataset, and their corresponding Radon-CDT representation is depicted in Figure \ref{fig:res2} (a) and (b). Figure \ref{fig:res2}, (c) and (d), show the projection of the data and the transformed data onto the top two {\it p}LDA directions learned from the data in the image space and in the Radon-CDT space for the face and the nuclei datasets, respectively (the images for the Radon transform and the Ridgelet transform are omitted for the sake of brevity). From both Figures \ref{fig:res01} and \ref{fig:res2}  it can be clearly seen that the Radon-CDT captures the nonlinearity of the data and simplifies the data structure significantly. 

\begin{figure}[t]
\centering
\includegraphics[width=.9\linewidth]{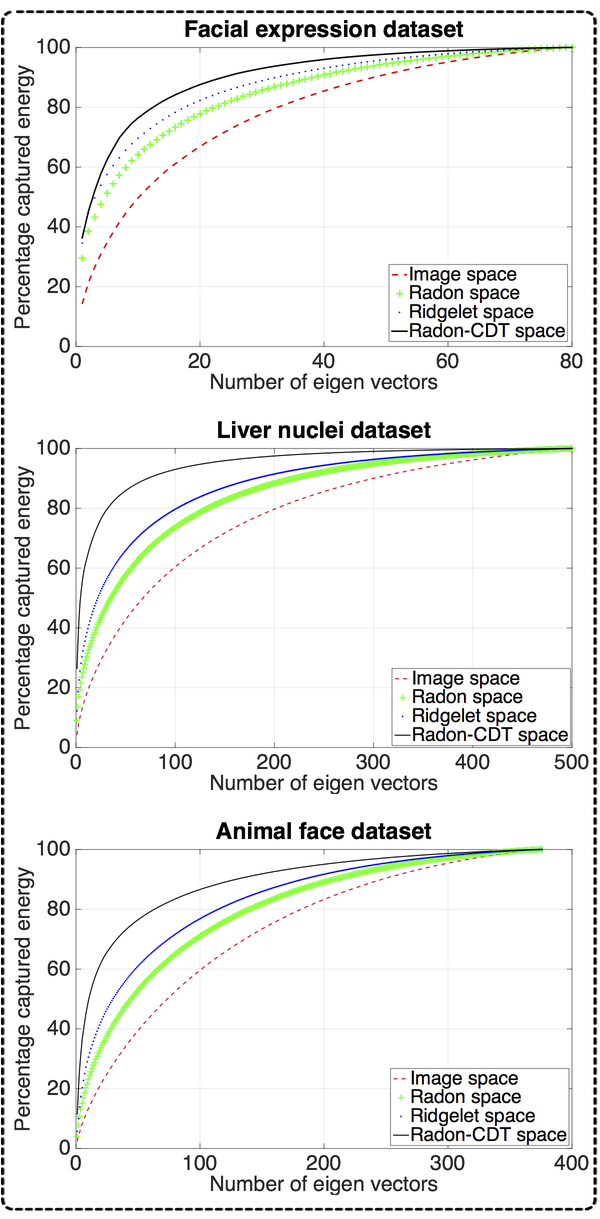}
\caption{Percentage variations captured by the principal components in the facial expression dataset (top) and the liver dataset (bottom), in the image space and in the Radon-CDT space.}
\label{fig:perE}
\end{figure}

 Note that, the eigenvalues of the covariance matrix of the data represent the amount of variations in the data that is captured by the principal components (i.e. its eigenvectors). Figure \ref{fig:perE} shows the cumulative percent variance (CPV) captured by the principal components calculated from the image space, the Radon transform space, the Ridgelet transform space, and the Radon-CDT space as a function of the number of principal components for all the datasets. It can be seen that the variations in the datasets are captured more efficiently and with fewer principal components in the Radon-CDT space as compared to the other transformation spaces. This indicates that the data structure becomes simpler in the Radon-CDT space, and the variations in the datasets can be explained with fewer parameters. 

We used the aforementioned datasets in supervised learning settings. The principal components of the datasets were first calculated and the data points were projected to these principal components (i.e. the dimensions which are not populated by data points were discarded). Next, a ten-fold cross validation scheme was used, in which $90\%$ of the data was used for training and the remaining $10\%$ was used for testing. A linear SVM classifier is learned and cross-validated  on the training data and the classification accuracy is calculated on the testing data. The average accuracy, averaged over the accuracies reported in the cross validation, for each dataset is reported in Table \ref{tab:res}. It can be seen that the linear classification accuracy is not only higher in the Radon-CDT space but also it is more consistent as the standard deviations of the reported accuracies are lower for the Radon-CDT space. We emphasize here that the use of linear SVM over kernel SVM, or any other nonlinear classifier (e.g. K nearest neighbors or random forest classifiers), is intentional. The classification experiments in this section serve as a measure of linear separability of image classes in the corresponding transform spaces and are designed to test our theorem on the linear separability of image classes in the Radon-CDT space. We note that one can utilize any preferred classifier or regressor in the Radon-CDT space.

To provide the reader with a reference point for comparing the classification results presented in Table \ref{tab:res}, we utilized the PCANet framework \cite{chan2014pcanet}, which is among the state of the art feature extraction methods, and applied it to our datasets. The extracted features from the PCANet are then used for classification. We emphasize that unlike the Radon-CDT, PCANet is not an invertible image transform and it only serves as a nonlinear feature extraction method. For PCANet we used $2$ layers with $8$ filters (principal patches) learned from the training data at each layer as suggested in \cite{chan2014pcanet}. The classification accuracy of PCANet is compared to that of the Radon-CDT in Table \ref{tab:pcanet}. It is clear the classification accuracies for all datasets are comparable. Here we need to note that changing the structure of the PCANet and fine-tuning it may lead to higher classification accuracies, but we utilized the parameters suggested in Chan et al.\cite{chan2014pcanet}.

\begin{figure}[t]
\centering
\includegraphics[width=.95\linewidth]{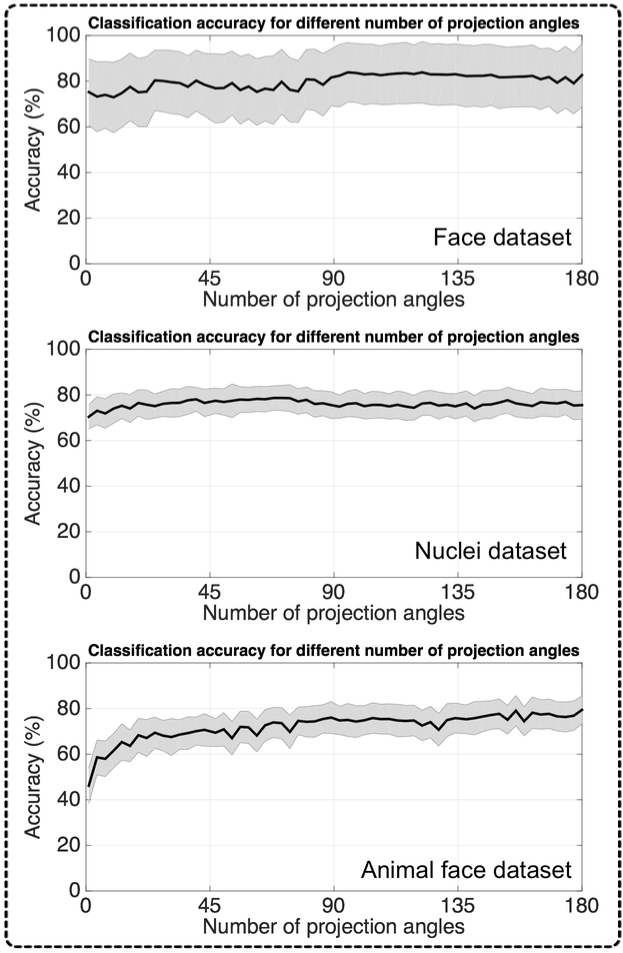}
\caption{Classification accuracy in the Radon-CDT space as a function of the number of projection angles.}
\label{fig:projectionAccuracy}
\end{figure}

In the experiments presented in this section we used 180 projection angles, $\theta\in [0^\circ,1^\circ,...,179^\circ]$. Here, a natural question arises regarding the dependency of the classification accuracies with the number of projection angles. To address this question, we experimentally tested the classification accuracies as a function of the number of projection angles. Figure \ref{fig:projectionAccuracy} shows the 10-fold cross validated mean accuracies with their corresponding standard deviations as a function of the number of projection angles for all datasets. From Figure \ref{fig:projectionAccuracy} it can be noticed that the classification accuracies are stable and there is a slight improvement in the accuracy as the number of projections increases. 

{\scriptsize
\begin{table}[t]
\centering
  \begin{tabular}{| l | c | c |}
    \hline
   \multirow{2}{*}{Face data} & \multicolumn{2}{ |c| }{Linear SVM } \\
  \cline{2-3}
 & Training accuracy & Testing Accuracy \\    
    \hline
	Image space& $100$ & $76.0\pm11.94$\\
    \hline
        Radon  space & $100$ & $79.12\pm12.25$	\\
       \hline
       Ridgelet  space & $100$ & $76.87\pm 13.95$\\
       \hline
	Radon-CDT space & $100$ & $82.62\pm11.5$\\
    \hline  
  \end{tabular}
  (a)
  
  \vspace{-.15in}
\begin{tabular}{c}  
\\
\end{tabular}

\begin{tabular}{| l | c | c |}
    \hline
   \multirow{2}{*}{Nuclei data} & \multicolumn{2}{ |c| }{Linear SVM } \\
  \cline{2-3}
 & Training accuracy & Testing Accuracy \\    
    \hline
    Image space& $100$ & $65.2\pm 6.6$\\    
    \hline
	Radon  space & $100$ & $62.56\pm6.7$	\\
       \hline
       Ridgelet  space & $100$ & $62.92\pm 5.6$\\
       \hline
    Radon-CDT space & $100$ & $75.56\pm6.21$\\
    \hline
\end{tabular}
(b)

\vspace{-.15in}
  \begin{tabular}{c}  
\\
\end{tabular}

\begin{tabular}{| l | c | c |}
    \hline
   \multirow{2}{*}{Animal Face data } & \multicolumn{2}{ |c| }{Linear SVM } \\
  \cline{2-3}
& Training accuracy & Testing Accuracy \\    
    \hline
    Image space& $100$ & $46.60\pm 7.71$\\    
    \hline
	Radon  space & $100$ & $47.94\pm 8.15$	\\
       \hline
       Ridgelet  space & $100$ & $69.39\pm7.07$\\
       \hline
    Radon-CDT space & $100$ & $79.42\pm6.12$\\
    \hline
  \end{tabular}
    (c)
    
  \caption{Average classification accuracy for the face dataset (a), the nuclei dataset (b), and the animal face dataset (c), calculated from ten-fold cross validation using linear SVM in the image space, the Radon transform space, the Ridgelet transform space, and the Radon-CDT spaces. The improvements are statistically significant for all datasets. }
\label{tab:res}
\end{table}}

{\scriptsize
\begin{table}[t]
\centering
\begin{tabular}{| l | c | c |}
    \hline
   \multirow{2}{*}{Classification comparison } & \multicolumn{2}{ |c| }{Linear SVM } \\
  \cline{2-3}
& PCANet & Radon-CDT \\    
    \hline
    Face dataset& $84.12\pm11.7$ & $82.62\pm11.5$\\    
    \hline
    Nuclei dataset & $74.16\pm 4.36$ & $75.56\pm6.21$	\\
       \hline
Animal face dataset & $80.81\pm5.1$ & $79.42\pm6.12$\\
    \hline
  \end{tabular}
    
  \caption{Average classification accuracy for the face dataset, the nuclei dataset, and the animal face dataset, calculated from ten-fold cross validation using linear SVM in the PCANet feature space and the Radon-CDT spaces. }
\label{tab:pcanet}
\end{table}}

\section{Summary and discussion}
\label{sec:con}

Problems involving classification of image data are pervasive in science and technology. Applications include automating and enhancing cancer detection from microscopic images of cells, person identification from images of irises or faces, mapping and identification of galaxy types from telescope images, and numerous others. The standard processing pipeline in these applications include 1) an image representation step, 2) feature extraction, and 3) statistical learning, though more recently, deep learning architectures have also been employed \cite{bengio2009learning} as an end to end learning approach. Regardless of the learning architecture being used, the image representation step is fundamental given that all feature extraction methods (e.g. SIFT, HOG, Haralick, etc. \cite{li2015feature}) require access to a representation model of pixel intensities. Many widely used mathematical image representation methods (e.g. wavelets, short time Fourier transforms, ridgelets, etc.) are linear, and thus, by themselves, are unable to enhance linear class separation in any way. Because linear operations are only able to analyze pixel intensities at fixed locations, they are unable to decode pixel displacements, which we hypothesize are crucial for better modeling intensity variations present in many classes of images. 

Here we described a new, non-linear, low-level image transform mean to exploit the hypothesis that analyzing pixel locations, in addition to their intensities, could be useful in problems of telling image classes apart. The new transform, termed the Radon-CDT transform, is derived by combining the 2D Radon transform \cite{quinto2006introduction} with the cumulative distribution transform (CDT) \cite{serim2015} described earlier. The transform is invertible as it contains well-defined forward (analysis) and inverse (synthesis) operations. We presented theoretical and experimental evidence towards supporting the hypothesis that the Radon-CDT can enhance the linear separability of certain signal classes. Underlying the theory is a specific generative model for signal classes which is non linear, and generates signals by transporting pixel intensities relative to a `mother' function. The theory and experimental results here add to our understanding in explaining why transport-based approaches have been able to improve the state of the art in certain cancer detection from microscopy images problems \cite{ozolek2014accurate, tosun2015detection}. 

In contrast to our earlier work related to transport-based signal and image analysis \cite{wang2013linear, wang2011optimal, basu2014detecting, serim2015, kolouri2015transport}, the work described here provides a number of important additions and improvements. First, in contrast to our earlier work for 1D signals \cite{serim2015}, the work presented here expands the concept of the CDT to 2D signals. In contrast to our earlier work in image analysis \cite{wang2013linear, basu2014detecting}, the Radon-CDT has a closed form, and hence does not require numerical optimization for computation. Because of this, theoretically analyzing certain of its properties with respect to image translation, scaling, and linear separability, becomes tractable (this analysis is presented in Section \ref{sec:Radon-CDT}). Finally, because the 2D Radon-CDT is closed form, it is also significantly faster, and simpler to compute.

Using an analogy to the kernel-based methods in machine learning, the Radon-CDT can be described as a kernel embedding space.  This connection between the Radon-CDT framework and the kernel methods is fully discussed in a recent work by the authors  \cite{kolouri2015sliced}. In short, we showed in \cite{kolouri2015sliced} that the kernel methods can be applied to both the image space and the Radon-CDT space, and demonstrated that applying the kernel methods in the Radon-CDT space lead to higher performance compared to applying them to the image space. In addition, as opposed to common kernel-based methods, in the Radon-CDT the transformation to the kernel space is known, and it is invertible at any point. This implies that
any statistical analysis in the Radon-CDT space can be `inverted' and presented back into the image space. We also presented theoretical results on the linear separability of data classes in this kernel-space (i.e. the Radon-CDT space) and how they are related to an image generative model which in addition to modifying pixel intensities, also displaces them in relation to a mother (template) image. The model suggests that pixel location information encoded in transport flows represents valuable information for simplifying classification problems. The model also allows one to potentially utilize any known physical information regarding the problem at hand (i.e. are classes expected to include translation, scaling, etc.) in considering whether the Radon-CDT would be an effective tool for solving it.

Finally, we note that, although transport-based methods, by themselves can at times improve upon state of the art methods in certain applications \cite{ozolek2014accurate, tosun2015detection, kolouri2015transport}, given its ability to simplify recognition tasks, we envision the Radon-CDT to serve as a low level pre-processing tool in classification problems. We note that because the proposed transform is mathematically invertible, it does not involve information loss. Thus, other representation methods such as wavelets, ridgelets, etc., as well as feature extraction methods (e.g. Haralick textures, etc.) can be employed in Radon-CDT space. Future work will involve designing numerically exact \emph{digital} versions of the image transformation framework presented here, as well as applying combinations of these techniques to numerous estimation and detection problems in image analysis and computer vision.

\section{Acknowledgement}
This work was financially supported in part by the National Science Foundation (NSF), grant number 1421502, the National Institutes of Health, grants   GM090033, CA188938, and GM103712, and the John and Claire Bertucci Graduate Fellowship.

\bibliographystyle{IEEE}
\bibliography{TIP_Radon_arXiv}

\section{Appendix}
\subsection{The Radon-CDT metric}
\vspace{-.05in}
\label{sec:appendix}
Here we show that $d_{RCD}(.,.)$ is indeed a metric as it satisfies, 
\begin{enumerate}[i.]
\item $d_{RCD}(I_1,I_0)\geq 0$
\begin{proof}
\begin{eqnarray*}
\Ih_0(t,\theta)> 0,~(f_1(t,\theta)-t)^2\geq 0, ~~~\forall\theta\in[0,\pi]\\ \Rightarrow d_{RCD}(I_1,I_0)\geq 0
\end{eqnarray*}
\end{proof}
\vspace{-.15in}
\item $d_{RCD}(I_1,I_0)= 0 \iff I_1=I_0$
\begin{proof}
\begin{eqnarray*}
d_{RCD}(I_1,I_0)&=&0 \iff \\ 
\int_{-\infty}^{\infty} (f_1(t,\theta)-t)^2\Ih_0(t,\theta) dt&=&0 ,~\forall\theta\in[0,\pi] \iff\\
(f_1(t,\theta)-t)^2&=&0,~\forall\theta\in[0,\pi] \iff\\
f_1(t,\theta)&=&t,~\forall\theta\in[0,\pi]  \iff\\
 \Ih_1(t,\theta)&=&\Ih_0(t,\theta), ~\forall\theta\in[0,\pi]    \\
\iff I_1(x,y)&=&I_0(x,y) 
\end{eqnarray*}\vspace{-.1in}\end{proof}
\vspace{-.15in}
\item $d_{RCD}(I_1,I_0)=d_{RCD}(I_0,I_1)$
\begin{proof}\begin{eqnarray*}
\vspace{-.2in}
d_{RCD}(I_1,I_0)= (\int_0^\pi \int_{-\infty}^{\infty} (f_1(t,\theta)-t)^2\Ih_0(t,\theta) dtd\theta)^{\frac{1}{2}}\\
= (\int_0^\pi \int_{-\infty}^{\infty} (u-f_1^{-1}(u,\theta))^2\frac{\partial f_1^{-1}}{\partial u} (u,\theta)
\\~~~~~~~~~~~~ \Ih_0(f_1^{-1}(u,\theta),\theta) dud\theta)^{\frac{1}{2}}\\
=(\int_0^\pi \int_{-\infty}^{\infty} (f^{-1}_1(u,\theta)-u)^2\Ih_1(u,\theta) dud\theta)^{\frac{1}{2}}\\
=d_{RCD}(I_0,I_1)
\end{eqnarray*}
where in the second line we used the change of variable $u=f(t,\theta)$ and in the fourth line we used $\frac{\partial f_1}{\partial t}(t,\theta)\Ih_1(f_1(t,\theta),\theta)=\Ih_0(t,\theta) \iff \frac{\partial f^{-1}_1}{\partial u}(u,\theta)\Ih_0(f^{-1}_1(u,\theta),\theta)=\Ih_1(u,\theta)$.\end{proof}
\item $d_{RCD}(I_1,I_2)\leq d_{RCD}(I_1,I_0)+d_{RCD}(I_2,I_0)$ 
\begin{proof}
let $\mu$, $\nu$, and $\sigma$ be the continuous probability measures on $\R^2$, with corresponding positive probability density functions $I_1$, $I_2$, and $I_0$. Let,
{\small
\begin{eqnarray*}
d^2_{RCD}(I_1,I_0)=\int_0^\pi \int_{-\infty}^{\infty} (f_1(t,\theta)-t)^2\Ih_0(t,\theta) dtd\theta\\
d^2_{RCD}(I_2,I_0)=\int_0^\pi \int_{-\infty}^{\infty} (f_2(t,\theta)-t)^2\Ih_0(t,\theta) dtd\theta\\
d^2_{RCD}(I_2,I_1)=\int_0^\pi \int_{-\infty}^{\infty} (h(t,\theta)-t)^2\Ih_1(t,\theta) dtd\theta.
\end{eqnarray*}}
Then we can write
\begin{eqnarray*}
d^2_{RCD}(I_2,I_1)=\int_0^\pi \int_{-\infty}^{\infty} (h(t,\theta)-t)^2\Ih_1(t,\theta) dtd\theta\\
=\int_0^\pi \int_{-\infty}^{\infty} (h(f_1(u,\theta),\theta)-f_1(u,\theta))^2\frac{\partial f_1}{\partial u}(u,\theta)\\~~~~~~~~~~~~~~\Ih_1(t,\theta) dud\theta \\
=\int_0^\pi \int_{-\infty}^{\infty} (f_2(u,\theta)-f_1(u,\theta))^2\Ih_0(t,\theta) dud\theta 
\end{eqnarray*}
where in the second line we used the change of variables $f_1(u,\theta)=t$. Defining $\f_i(\brho)=[f_i(t,\theta),\theta]^T$ for $i=1,2$, where $\brho=[t,\theta]^T$, above equation can be written as a weighted Euclidean distance with weights $\Ih_0(\brho)$. Therefore we can write, 
\begin{eqnarray*}
d_{RCD}(I_2,I_1)
&=&\| (\f_1-id)-(\f_2-id)\|_\sigma\\
&\leq&\| \f_1-id\|_\sigma+\|\f_2-id\|_\sigma\\
&=& d_{RCD}(I_1,I_0)+d_{RCD}(I_2,I_0)
\end{eqnarray*}
\end{proof}
\vspace{-.2in}
\end{enumerate}
\vspace{-.2in}
\subsection{Translation property of Radon-CDT}
\vspace{-.05in}
\label{sec:ptrans}
For $J(x,y)=I(x-x_0,y-y_0)$ and using the properties of Radon transform we have, 
\begin{eqnarray*}
\hat{J}(t,\theta)=\Ih(t-x_0\cos(\theta)-y_0\sin(\theta),\theta)
\end{eqnarray*}
Therefore the Radon-CDT of $J$ can be written as, 
\begin{eqnarray*}
\tilde{J}(t,\theta)= (g(t,\theta)-t)\sqrt{\Ih_0(t,\theta)}
\end{eqnarray*}
where $g(t,\theta)$ satisfies,
\begin{eqnarray*}
\int_{-\infty}^{g(t,\theta)} \hat{J}(\tau,\theta)d\tau= \int_{-\infty}^{t}\Ih_0(\tau,\theta)d\tau
\end{eqnarray*}
The left hand side of above equation can be rewritten as, 
\begin{eqnarray*}
\int_{-\infty}^{g(t,\theta)} \hat{J}(\tau,\theta)d\tau= \int_{-\infty}^{g(t,\theta)}\Ih(\tau-x_0\cos(\theta)-y_0\sin(\theta),\theta)d\tau\\
= \int_{-\infty}^{g(t,\theta)-x_0\cos(\theta)-y_0\sin(\theta)}\Ih(u,\theta)du \Rightarrow\\
g(t,\theta)-x_0\cos(\theta)-y_0\sin(\theta)=f(t,\theta) \Rightarrow\\
g(t,\theta)=f(t,\theta)+x_0\cos(\theta)+y_0\sin(\theta) \Rightarrow\\
(g(t,\theta)-t)\sqrt{\Ih_0(t,\theta)}=(f(t,\theta)-t)\sqrt{\Ih_0(t,\theta)}+\\(x_0\cos(\theta)+y_0\sin(\theta))\sqrt{\Ih_0(t,\theta)}\Rightarrow\\
\tilde{J}(t,\theta)=\It(t,\theta)+(x_0\cos(\theta)+y_0\sin(\theta))\sqrt{\Ih_0(t,\theta)}
\end{eqnarray*}
where $\frac{\partial f}{\partial t}(t,\theta)\Ih(f(t,\theta),\theta)=\Ih_0(t,\theta)$.

\subsection{Scaling property of Radon-CDT}
\vspace{-.05in}
\label{sec:pscale}
For $J(x,y)=\alpha^2I(\alpha x,\alpha y)$ with $\alpha>0$ and using the properties of Radon transform we have, 
\begin{eqnarray*}
\hat{J}(t,\theta)=\alpha \Ih(\alpha t,\theta).
\end{eqnarray*}
The Radon-CDT of $J$ can be written as, 
\begin{eqnarray*}
\tilde{J}(t,\theta)= (g(t,\theta)-t)\sqrt{\Ih_0(t,\theta)}
\end{eqnarray*}
where $g(t,\theta)$ satisfies,
\begin{eqnarray*}
\int_{-\infty}^{g(t,\theta)} \hat{J}(\tau,\theta)d\tau= \int_{-\infty}^{t}\Ih_0(\tau,\theta)d\tau
\end{eqnarray*}
The left hand side of above equation can be rewritten as, 
\begin{eqnarray*}
\int_{-\infty}^{g(t,\theta)} \hat{J}(\tau,\theta)d\tau= \int_{-\infty}^{g(t,\theta)}\alpha \Ih(\alpha\tau,\theta)d\tau\\
=\int_{-\infty}^{\alpha g(t,\theta)}\Ih(u,\theta)du \Rightarrow\\
g(t,\theta)=\frac{f(t,\theta)}{\alpha} \Rightarrow\\
(g(t,\theta)-t)\sqrt{\Ih_0(t,\theta)}=(\frac{f(t,\theta)}{\alpha}-t)\sqrt{\Ih_0(t,\theta)}\\
=\frac{(f(t,\theta)-t)\sqrt{\Ih_0(t,\theta)}}{\alpha}+(\frac{1-\alpha}{\alpha})\sqrt{\Ih_0(t,\theta)}\Rightarrow\\
\tilde{J}(t,\theta)=\frac{\It(t,\theta)}{\alpha}+ (\frac{1-\alpha}{\alpha})\sqrt{\Ih_0(t,\theta)}
\end{eqnarray*}

\subsection{Rotation property of Radon-CDT}
\label{sec:prot}
For $J(x,y)=I(x\cos(\phi)+y\sin(\phi),-x\sin(\phi)+y\cos(\theta))$ and using the properties of Radon transform we have, 
\begin{eqnarray*}
\hat{J}(t,\theta)=\Ih(t,\theta-\phi).
\end{eqnarray*}
Given a circularly symmetric reference image, the Radon-CDT of $J$ can be written as, 
\begin{eqnarray*}
\tilde{J}(t,\theta)= (g(t,\theta)-t)\sqrt{\Ih_0(t,\theta)}
\end{eqnarray*}
where $g(t,\theta)$ satisfies,
\begin{eqnarray*}
\int_{-\infty}^{g(t,\theta)} \hat{J}(\tau,\theta)d\tau= \int_{-\infty}^{t}\Ih_0(\tau,\theta)d\tau
\end{eqnarray*}
The left hand side of above equation can be rewritten as, 
\begin{eqnarray*}
\int_{-\infty}^{g(t,\theta)} \hat{J}(\tau,\theta)d\tau= \int_{-\infty}^{g(t,\theta+\phi)} \Ih(\tau,\theta)d\tau \Rightarrow\\
f(t,\theta)=g(t,\theta+\phi) \Rightarrow g(t,\theta)=f(t,\theta-\phi)\Rightarrow\\
(g(t,\theta)-t)\sqrt{\Ih_0(t,\theta)}=(f(t,\theta-\phi)-t)\sqrt{\Ih_0(t,\theta)}\\
=(f(t,\theta-\phi)-t)\sqrt{\Ih_0(t,\theta-\phi)} \Rightarrow\\
\tilde{J}(t,\theta)=\It(t,\theta-\phi)
\end{eqnarray*}

\subsection{Linear separability in the Radon-CDT space}
\label{sec:separ}
Let image classes $\P$ and $\Q$ be generated from Eq. \eqref{eq:PQ}. Here we show that the classes are linearly separable in the Radon-CDT space. 
\begin{proof}
By contradiction we assume that the transformed image classes are not linearly separable,
\begin{eqnarray*}
\sum_i \alpha_i \tilde{p}_i(t,\theta)&=&\sum_j \beta_j \tilde{q}_j(t,\theta)  \Rightarrow\\
\sum_i \alpha_i f_i(t,\theta) &=&\sum_j \beta_j g_j(t,\theta)
\end{eqnarray*}
where $\sum_i \alpha_i=\sum_j \alpha_j=1$,  $\frac{\partial f_i}{\partial t}(t,\theta)\hat{p}_i(f_i( t,\theta),\theta)=\Ih_0(t,\theta)$, and $\frac{\partial g_j}{\partial t}(t,\theta)\hat{q}_j(g_j( t,\theta),\theta)=\Ih_0(t,\theta)$. Figure \ref{fig:diagram} shows a diagram which illustrates the interactions between $\hat{q}_j$s, $\hat{p}_i$s, and  $\Ih_0$. 
\begin{figure}[t]
\centering
\includegraphics[width=.45\columnwidth]{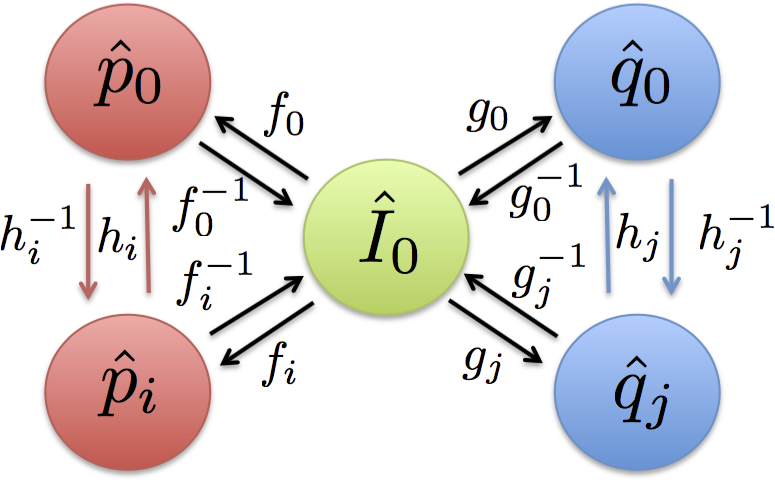}
\caption{The diagram of interactions of the images mass preserving maps.}
\label{fig:diagram}
\end{figure}
It is straightforward to show that $f_i(t,\theta)=h^{-1}_i(f_0(t,\theta),\theta)$  and $g_j(t,\theta)=h^{-1}_j(g_0(t,\theta),\theta)$ as  can also be seen from the diagram in Figure \ref{fig:diagram}. Therefore we can write,
\begin{eqnarray*}
\sum_i \alpha_i f_i(t,\theta) &=&\sum_j \beta_j g_j(t,\theta)\Rightarrow\\
\sum_i \alpha_i h^{-1}_i(f_0(t,\theta),\theta)&=&\sum_j \beta_j h^{-1}_j(g_0(t,\theta),\theta)
\end{eqnarray*}
Defining $h_\alpha(t,\theta)= \sum_i \alpha_i h^{-1}_i(t,\theta)\in \C$ and  $h_\beta(t,\theta)=\sum_i \beta_j h^{-1}_j(t,\theta)\in \C$, we can rewrite above equation as,
\begin{eqnarray*}
h_\alpha(f_0(t,\theta),\theta)&=&h_\beta(g_0(t,\theta),\theta)\Rightarrow\\
f_0(t,\theta)&=& h^{-1}_\alpha(h_\beta(g_0(t,\theta),\theta),\theta).
\end{eqnarray*}
Defining $h(t,\theta)=h^{-1}_\alpha(h_\beta(t,\theta),\theta)\in \C$ we have, 
\begin{eqnarray*}
f_0(t,\theta)= h(g_0(t,\theta),\theta) 
\end{eqnarray*}
which implies that $\exists h \in \C \rightarrow \frac{\partial h}{\partial t}(t,\theta)\hat{p}_0(h(t,\theta),\theta)=\hat{q}_0(t,\theta)$, which contradicts with the fourth condition of $\C$. 

\end{proof}

\end{document}